\documentclass[runningheads]{llncs}
\usepackage[T1]{fontenc}
\usepackage{graphicx}

\PassOptionsToPackage{numbers, compress}{natbib}
\usepackage[utf8]{inputenc} 
\usepackage[T1]{fontenc}    
\usepackage{hyperref}       
\usepackage{url}            
\usepackage{booktabs}       
\usepackage{amsfonts}       
\usepackage{nicefrac}       
\usepackage{microtype}      
\usepackage{algorithm}
\usepackage{mathtools}
\usepackage{placeins}
\usepackage{algpseudocode}

\usepackage{bm}
\usepackage{rotating}
\usepackage{graphicx}
\usepackage{float}
\usepackage{xcolor}

\begin{document}
%

\title{HabitatAgent: An End-to-End Multi-Agent System for Housing Consultation}
%
%

\author{Hongyang Yang\inst{1}\thanks{Corresponding author: \email{hy2500@columbia.edu}. Hongyang Yang serves as the CTO of Fangdongdong.} \and
Yanxin Zhang\inst{1} \and
Yang She\inst{3}  \and
Yue Xiao\inst{2}  \and
Hao Wu\inst{1}  \and
Yiyang Zhang\inst{1}  \and
Jiapeng Hou\inst{1}  \and
Rongshan Zhang\inst{1}
}
\authorrunning{H. Yang et al.}
%
\institute{Fangdongdong,  \and
Tsinghua University,  \and
Columbia University}
%
\maketitle              
 \vspace{-6mm}
\begin{abstract}
Housing selection is a high-stakes and largely irreversible decision problem. We study \emph{housing consultation} as a decision-support interface for housing selection. Existing housing platforms and many LLM-based assistants often reduce this process to ranking or recommendation, resulting in opaque reasoning, brittle multi-constraint handling, and limited guarantees on factuality.

We present \textbf{HabitatAgent}, the first LLM-powered multi-agent architecture for end-to-end housing consultation. HabitatAgent comprises four specialized \textbf{agent roles}: \textbf{Memory}, \textbf{Retrieval}, \textbf{Generation}, and \textbf{Validation}. The \textbf{Memory Agent} maintains multi-layer user memory through internal stages for constraint extraction, memory fusion, and verification-gated updates; the \textbf{Retrieval Agent} performs \textit{hybrid vector--graph retrieval} (GraphRAG); the \textbf{Generation Agent} produces evidence-referenced recommendations and explanations; and the \textbf{Validation Agent} applies multi-tier verification and targeted remediation. Together, these agents provide an auditable and reliable workflow for end-to-end housing consultation.

We evaluate HabitatAgent on \textbf{100 real user consultation scenarios} (\textbf{300} multi-turn question--answer pairs) under an \emph{end-to-end correctness} protocol. A strong single-stage baseline (Dense+Rerank) achieves \textbf{75\%} accuracy, while HabitatAgent reaches \textbf{95\%}.
\end{abstract}
 \vspace{-6mm}

\section{Introduction}

Housing selection is a consequential and largely irreversible decision with high switching costs and strong path dependence. Unlike many reversible consumer choices, it involves substantial transaction frictions, limited liquidity, and pervasive information asymmetry. Homebuyers therefore must reason over heterogeneous and sometimes conflicting factors—including budget, location, accessibility, building quality, community services, and policy constraints—often under incomplete or noisy evidence.


We study \emph{housing consultation} as a decision-support interface for housing selection, whereas most existing platforms still reduce it to \emph{search and ranking}~\cite{gharahighehi2021realestate_recsys_survey,henriquezmiranda2025realestate_systematic_review,oberoi2024re,ball1994using}. Conversational recommendation, LLM-based assistants, and multi-agent systems have improved multi-turn interaction, preference modeling, and task decomposition~\cite{jannach2022conversational_rec_grand_challenge,zhu2025crave,li2024llm_multiagent_survey,wang2024macrec,yang2023fingpt_open}. However, reliable housing consultation remains challenging in practice because many systems still rely on monolithic prompting, weak evidence grounding, and limited safeguards against factual and entity errors.

When framed as an end-to-end housing consultation problem, these limitations manifest as four concrete challenges:
\begin{enumerate}
  \item \textbf{Evolving and under-specified user preferences.}
  Buyers often begin with vague goals and progressively refine constraints over multi-turn dialogue; the system must elicit latent needs, disambiguate intent, and maintain a consistent preference state across turns~\cite{jannach2022conversational_rec_grand_challenge,zhu2025crave}.

  \item \textbf{Heterogeneous evidence and relational constraints.}
  Accurate consultation requires integrating heterogeneous sources (projects, transit, schools, policies, costs) and enforcing relational or hard constraints that cannot be satisfied by semantic similarity alone, motivating graph-aware retrieval~\cite{han2025graphrag,microsoft_graphrag_github}.

  \item \textbf{Opaque recommendations and ungrounded shortlist decisions.}
  Users need auditable comparisons and explicit rationales aligned with their priorities, rather than opaque top-$k$ ranking outputs~\cite{gharahighehi2021realestate_recsys_survey,henriquezmiranda2025realestate_systematic_review}.

  \item \textbf{High cost of factual and entity errors.}
    Misstated numbers, conflated entities, or hallucinated amenities can directly mislead irreversible decisions, requiring explicit verification and targeted correction~\cite{farquhar2024semantic_entropy}.
\end{enumerate}

To address these challenges, we propose \textbf{HabitatAgent}, a production-oriented multi-agent architecture for end-to-end housing consultation. HabitatAgent organizes four specialized roles---\textbf{Memory}, \textbf{Retrieval}, \textbf{Generation}, and \textbf{Validation}---into a closed-loop workflow for reliable decision support. The system is built on three key mechanisms: (i) \emph{Verification-Gated Memory}, which prevents unverified information from contaminating long-term user state; (ii) \emph{Adaptive Retrieval Routing}, which selectively invokes graph-constrained retrieval for relationally complex queries; and (iii) \emph{Failure-Type-Aware Remediation}, which applies targeted recovery instead of naive regeneration when validation fails. On 100 real consultation scenarios (300 multi-turn Q\&A pairs), HabitatAgent improves end-to-end accuracy from 75\% (Dense+Rerank) to 95\%.

\paragraph{Contributions.}
This work makes three contributions:
\begin{enumerate}
    \item We formulate buyer-side housing consultation as an \emph{end-to-end, high-stakes decision-support problem}, beyond ranking-only real-estate recommendation.
    \item We present a closed-loop multi-agent architecture that couples \emph{verification-gated memory}, \emph{adaptive vector--graph retrieval routing}, and \emph{failure-type-aware remediation} for reliable multi-turn, multi-constraint consultation.
    \item We show on real consultation scenarios that this design substantially improves end-to-end correctness over strong dense-retrieval, graph-retrieval, and self-correction baselines.
\end{enumerate}

\section{Related Work}

\subsection{Conversational and Decision-Support Systems for Housing}
Housing consultation differs from standard real-estate recommendations because it involves multi-turn preference refinement, heterogeneous evidence, and high-stakes trade-offs. Prior work on real-estate recommender systems mainly studies ranking, filtering, and preference matching~\cite{henriquezmiranda2025realestate_systematic_review}. These systems are useful for candidate generation, but they usually treat the task as search or recommendation rather than end-to-end consultation, and therefore provide limited support for iterative clarification, explicit evidence grounding, and post-generation verification.

\subsection{LLMs and Agents in Real-Estate Applications}
Recent work has explored LLMs in real-estate settings, including listing-oriented generation, domain-adapted decision support, and agent-based assistants~\cite{wu2025ai,zhu2025real,haurum2024realestateai}. These studies show the promise of LLMs in the domain, but most focus on a single capability---such as generation, appraisal, or assistant interaction---rather than a closed-loop consultation workflow that jointly manages memory, retrieval, generation, and validation.


\subsection{Multi-Agent LLM Systems and Graph-Grounded Retrieval}
In recent years, large language models (LLMs) have been increasingly applied to specialized vertical industries, such as finance and healthcare. Examples include BloombergGPT~\cite{wu2023bloomberggpt}, FinGPT~\cite{yang2023fingpt_open,wang2023fingptbenchmark,liang2024fingpt}, and FinRobot~\cite{yang2024finrobot,zhou2024finrobot,han2024enhancing}, which leverage LLMs to assist in complex decision-making tasks in their respective domains. These systems highlight the potential of LLMs in providing tailored solutions, demonstrating how they can be employed for high-precision, domain-specific tasks. More broadly, LLM-based multi-agent systems have emerged as a general paradigm for decomposing complex tasks into specialized roles~\cite{li2024llm_multiagent_survey,tran2025collaboration_survey,guo2024llm_multiagents_ijcai}. These directions are highly relevant to housing consultation, where users often impose relational constraints such as transit access, school districts, or location dependencies. In addition, recent housing-related benchmarks and datasets further highlight the importance of grounded evidence and reliable decision support in this domain~\cite{bagalkotkar2024fairhome,zhu2025real}.

\subsection{Gap and Positioning of This Work}
Our work differs from prior studies in two ways. First, we focus on buyer-side housing consultation as an end-to-end decision-support problem rather than a ranking-only or generation-only task. Second, the main contribution is not a single isolated module, but a closed-loop architecture that couples three mechanisms: verification-gated memory updates, adaptive vector--graph retrieval routing, and failure-type-aware remediation. This combination is designed specifically for high-stakes, multi-constraint consultation settings where correctness, traceability, and recovery matter as much as recommendation relevance.

\section{Methodology}
\label{sec:methodology}

HabitatAgent is a multi-agent architecture for end-to-end housing consultation. It organizes four specialized agents---\textbf{Memory}, \textbf{Retrieval}, \textbf{Generation}, and \textbf{Validation}---into a closed-loop workflow for reliable decision support. The design objective is not merely task decomposition, but robust decision support under evolving preferences, relational constraints, and high error cost.

\begin{figure*}[t]
  \centering
  \makebox[\textwidth][c]{%
    \includegraphics[width=1.65\textwidth]{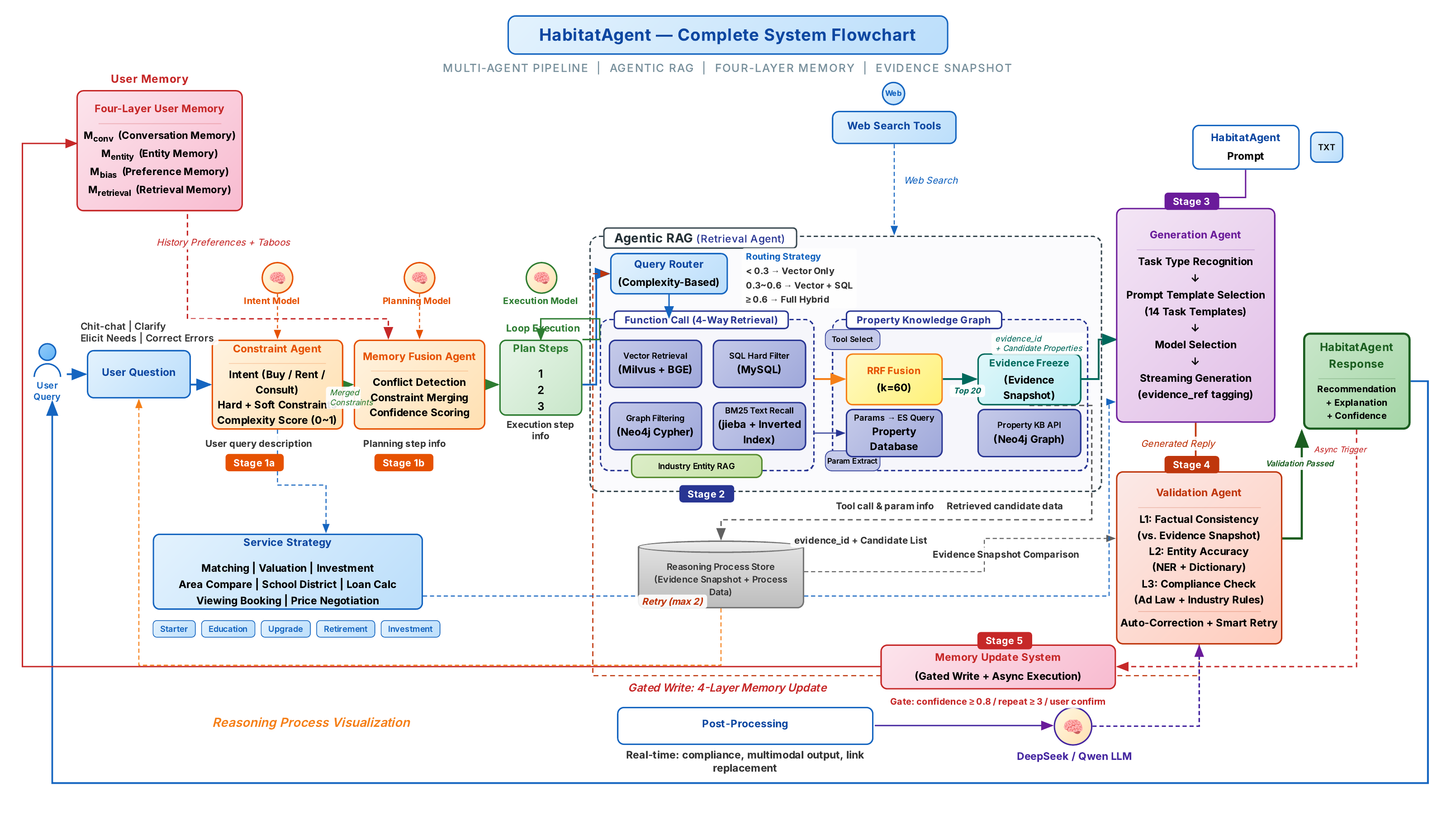}%
  }
  \vspace{-5mm}
  \caption{HabitatAgent overview. Four specialized agents---\textbf{Memory}, \textbf{Retrieval}, \textbf{Generation}, and \textbf{Validation}---are executed as a \textbf{five-stage} workflow. In particular, Stage~1a (constraint extraction), Stage~1b (memory fusion), and Stage~5 (memory update) are internal substeps of the \textbf{Memory Agent}.}
  \label{fig:architecture}
\end{figure*}



Figure~\ref{fig:architecture} summarizes the full workflow. At each turn, HabitatAgent first consolidates user state, then retrieves auditable evidence, generates a grounded response, and finally validates the output with targeted remediation when needed. The three methodological mechanisms studied in this paper---Verification-Gated Memory, Adaptive Retrieval Routing, and Failure-Type-Aware Remediation---are implemented across this workflow.

\begin{figure*}[t]
  \centering
  \makebox[\textwidth][c]{%
    \includegraphics[width=0.98\textwidth]{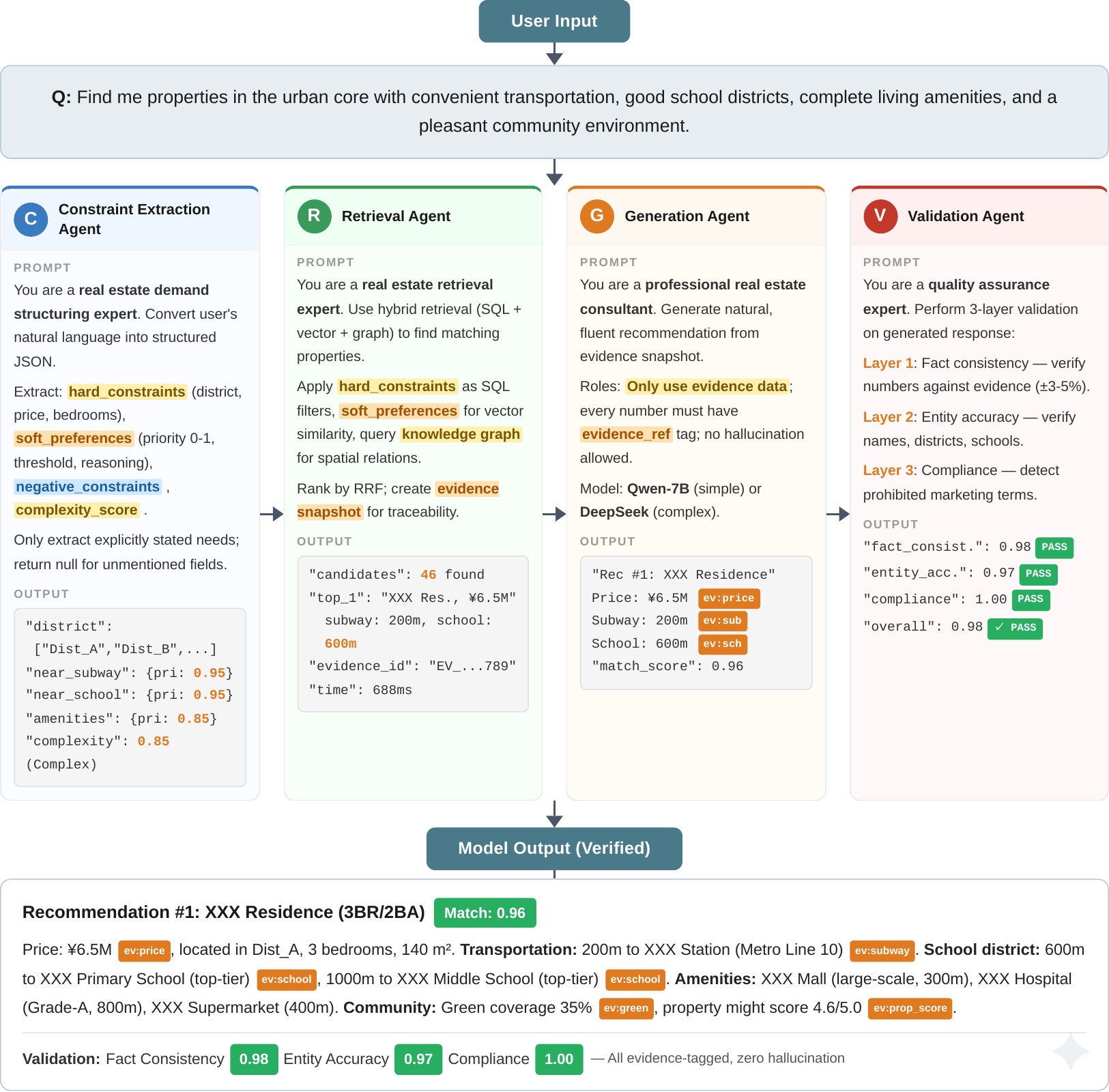}%
  }
   \vspace{-3mm}
  \caption{Use case of HabitatAgent. A user query is processed through constraint extraction, hybrid retrieval (GraphRAG), evidence-grounded generation, and multi-tier validation to produce a verified recommendation.}
  \label{fig:usecase}
\end{figure*}

Figure~\ref{fig:usecase} further illustrates a concrete end-to-end example of how HabitatAgent processes a realistic housing query into a verified recommendation.


 \vspace{-6mm}
 
\subsection{Memory Agent: Verification-Gated Multi-Layer Memory}
\label{subsec:memory_agent}
Multi-turn housing consultation requires persistent preference tracking, yet naive memory accumulation can cause long-term drift due to polluted or unverified information. HabitatAgent addresses this tension via a \textbf{Verification-Gated Memory} mechanism: only information extracted from responses that pass multi-tier validation is allowed to update long-term user memory. This design prevents error propagation across turns and stabilizes personalization over time.

\subsubsection{Four-layer memory structure}
The Memory Agent maintains a four-layer hierarchy:
\begin{itemize}
    \item \textbf{Conversational memory ($M_{\text{conv}}$)}: a short-term buffer storing the most recent 5 turns, with a TTL of 24 hours.
    \item \textbf{Entity memory ($M_{\text{entity}}$)}: a mid-term store of referenced entities (e.g., properties, regions) represented as a weighted sorted list.
    \item \textbf{Bias memory ($M_{\text{bias}}$)}: a long-term store of explicit preferences/aversions (e.g., ``dislikes noisy areas''), each with a weight $w\in[-1,1]$.
    \item \textbf{Retrieval memory ($M_{\text{retrieval}}$)}: a 7-day rolling cache of previously recommended property IDs, used for deduplication and diversification.
\end{itemize}

\paragraph{Toward context-aware conversational memory.}
Our current conversational memory $M_{\text{conv}}$ keeps the most recent 5 turns to control context length and latency. However, real housing consultation can involve long-horizon reference and backtracking (e.g., comparing a property mentioned much earlier). To mitigate context fragmentation, we optionally maintain a lightweight \textbf{context-aware state} that summarizes salient constraints and referenced entities, and updates this state at each turn (e.g., \textsc{Add}/\textsc{Update}/\textsc{Relax}). This allows the Memory Agent to recover important earlier context without retaining the full raw dialogue.

\subsubsection{Gated update policy}
Let $\text{extract}(Q_t,A_t)$ denote preference tuples extracted from the query--answer pair at turn $t$. The Memory Agent updates state only when the response $A_t$ is validated:
\begin{equation}
    M_{t+1} = 
    \begin{cases} 
        M_t \cup \{\text{extract}(Q_t, A_t)\}, & \text{if } V(A_t,R_t)=\text{pass} \\
        M_t, & \text{otherwise} 
    \end{cases}
\end{equation}
where $R_t$ denotes retrieved evidence used at turn $t$ (cf. \S\ref{subsec:retrieval_agent}). The primary validation function focuses on factual and entity correctness:
\begin{equation}
    V(A,R)=
    \begin{cases}
      \text{pass}, & \text{if } V_{\text{fact}}(A,R)\ge 0.85 \land V_{\text{entity}}(A,R)\ge 0.90 \\
      \text{fail}, & \text{otherwise}
    \end{cases}
\end{equation}
where $V_{\text{fact}}$ checks factual consistency (e.g., price, size) and $V_{\text{entity}}$ checks entity correctness (e.g., property names). In addition, we apply an auxiliary compliance filter $V_{\text{comp}}(A)$ as a conservative safety safeguard, but it is not part of the primary evaluation metric in \S\ref{sec:evaluation}.

\subsection{Retrieval Agent: Adaptive Hybrid Vector--Graph Retrieval (GraphRAG)}
\label{subsec:retrieval_agent}
Housing consultation requires evidence that is both semantically relevant and structurally consistent with hard/relational constraints (e.g., ``near Line~10'' or ``within 30 minutes to CBD''). Pure dense retrieval is fast but may violate relational constraints; pure graph filtering is precise but expensive. HabitatAgent therefore employs a \textbf{hybrid vector--graph retrieval} design, and introduces an \textbf{Adaptive Retrieval Router} to decide when to invoke graph-constrained retrieval.

\subsubsection{Adaptive retrieval router}
The router $f_{\theta}: Q \rightarrow [0,1]$ predicts whether a query requires graph retrieval. It consumes a feature vector $\phi(Q)$ including:
\begin{itemize}
    \item number of extracted constraints $N_c$,
    \item count of relational keywords (e.g., ``near'', ``commute to'') $N_r$,
    \item confidence drop in dense retrieval (e.g., $\text{score}_1-\text{score}_5$),
    \item a binary history flag indicating whether similar queries previously failed ($\text{fail}_h$).
\end{itemize}
We optimize $f_{\theta}$ with a cost-sensitive loss, penalizing false negatives ($C_{\text{FN}}=5$) more than false positives ($C_{\text{FP}}=1$), since failing to route complex queries to graph retrieval is more harmful than paying extra retrieval cost.

\subsubsection{Graph-constrained retrieval}
For queries routed as complex, the Retrieval Agent performs two-step hybrid retrieval: (i) dense retrieval returns a broad candidate set (top 100), then (ii) constraints are translated into a Cypher query executed over a property knowledge graph to filter candidates. Our graph contains 6{,}016 nodes (properties, regions, transit, schools, districts) and 45{,}000 edges encoding relations such as \texttt{LOCATED\_IN} and \texttt{NEAR\_SUBWAY}.

\subsection{Generation Agent: Task-Aware Prompted Response Generation}
\label{subsec:generation_agent}
Given structured constraints from the Memory Agent and evidence $R_t$ from the Retrieval Agent, the Generation Agent produces \textbf{context-aware} and \textbf{evidence-referenced} responses through a task-aware orchestration layer. Instead of relying on a single prompt, it first identifies the task type of the current query and then selects one of \textbf{14 task-specific prompt templates} covering common housing consultation scenarios, including recommendation, property query, comparison, facility query, value analysis, investment, school-district consultation, first-time buyer guidance, second-hand housing, decoration, out-of-town purchase, short-term rental, policy interpretation, and general fallback.

This design enables a unified generation interface across diverse consultation intents while preserving evidence grounding. The Generation Agent then selects the appropriate model and produces a response with explicit evidence references for key factual claims, improving transparency and supporting downstream validation (\S\ref{subsec:validation_agent}).

\subsection{Validation Agent: Multi-Tier Verification and Failure-Type-Aware Remediation}
\label{subsec:validation_agent}
Because factual and entity errors can directly mislead high-stakes decisions, HabitatAgent verifies each candidate response before presenting it to the user and before committing updates to long-term memory. The Validation Agent applies \textbf{multi-tier checks} and triggers \textbf{failure-type-aware remediation} to recover from common failure modes instead of terminating.

\subsubsection{Failure classification}
When validation fails, the agent labels the failure as one of:
\begin{itemize}
    \item \textbf{Entity missing}: the response mentions an entity not supported by retrieved evidence.
    \item \textbf{Constraint conflict}: retrieved results cannot satisfy constraints simultaneously.
    \item \textbf{Factual error}: numeric/categorical claims contradict evidence.
\end{itemize}

\subsubsection{Remediation policy}
Let $V_{\text{fail}}$ denote the failure type. The remediation policy maps failures to targeted actions:
\begin{equation}
\text{Remediate}(V_{\text{fail}})=
\begin{cases}
    \text{RetrieveByEntity}(e_{\text{miss}}), & \text{if entity\_missing} \\
    \text{RelaxThreshold}(\tau \rightarrow 0.9\tau), & \text{if constraint\_conflict} \\
    \text{LocalCorrect}(A, V_{\text{issues}}), & \text{if fact\_error}
\end{cases}
\end{equation}
Entity-missing triggers a retrieval centered on $e_{\text{miss}}$; constraint-conflict relaxes the least important retrieval threshold and retries; factual-error performs local correction guided by identified issues, avoiding full regeneration. This closed-loop design increases the rate of \emph{validated} responses and enables \textbf{verification-gated} memory updates in \S\ref{subsec:memory_agent}.

\section{Evaluation}
\label{sec:evaluation}

We evaluate HabitatAgent as an \emph{auditable, end-to-end} housing consultation system. The design of our experiments follows the system logic introduced in \S\ref{sec:methodology}: (i) \textbf{Memory} supports evolving preferences via verification-gated updates, (ii) \textbf{Retrieval} handles heterogeneous and relational constraints via adaptive hybrid vector--graph retrieval, (iii) \textbf{Generation} produces evidence-referenced recommendations and explanations, and (iv) \textbf{Validation} enforces factual/entity correctness with failure-type-aware remediation.

Our evaluation answers two research questions:
\begin{itemize}
    \item \textbf{RQ1 (Overall Effectiveness):} Does HabitatAgent improve \emph{end-to-end correctness} and recommendation quality over strong single-stage and graph-based baselines?
    \item \textbf{RQ2 (Component Contribution):} How much do \textbf{Verification-Gated Memory}, \textbf{Adaptive Retrieval Routing}, and \textbf{Failure-Type-Aware Remediation} each contribute to the final performance?
\end{itemize}

\subsection{Experimental Setup}
\label{subsec:eval_setup}

\subsubsection{Data}
We use a proprietary dataset derived from real, anonymized user interactions on a Beijing housing platform.
\begin{itemize}
    \item \textbf{Property corpus.} 5{,}000 property listings with structured attributes (e.g., price, layout, area, geo-coordinates) and linked amenities (e.g., subway, schools). This corpus is used to construct both the vector index and the property knowledge graph.
    \item \textbf{Consultation scenarios.} 100 real consultation scenarios, each with a 3-turn dialogue, resulting in 300 user queries for end-to-end evaluation. Queries are categorized as \textit{Simple} (80\%, 1--2 constraints) or \textit{Complex} (20\%, $\ge$3 constraints and/or relational requirements).
    \item \textbf{Human annotations.} Three trained annotators label each query-response pair for constraint satisfaction and factual/entity correctness. Inter-annotator agreement is substantial (Cohen's $\kappa=0.82$).
\end{itemize}
\textbf{Ethics.} All user data is anonymized and used with explicit consent for research purposes.
\begin{table}[t]
\centering
\caption{Overall performance comparison on the full dataset (300 queries). Best results are in \textbf{bold}.}
\label{tab:main_results}
\resizebox{\columnwidth}{!}{
\begin{tabular}{@{}lcccc@{}}
\toprule
\textbf{System} & \textbf{Accuracy} & \textbf{nDCG@5} & \textbf{Faithfulness} & \textbf{P95 Latency (ms)} \\
\midrule
B1: Monolithic RAG & 0.72 & 0.76 & 0.78 & 450 \\
B2: Dense+Rerank & 0.75 & 0.80 & 0.82 & 380 \\
B3: GraphRAG-Fixed & 0.82 & 0.85 & 0.88 & 820 \\
B4: LLM-Ranker & 0.70 & 0.78 & 0.75 & 1200 \\
B5: Self-RAG & 0.78 & 0.82 & 0.85 & 680 \\
B6: Rule-Verifier & 0.80 & 0.83 & 0.86 & 520 \\
\midrule
\textbf{Ours: HabitatAgent} & \textbf{0.95} & \textbf{0.92} & \textbf{0.96} & \textbf{720} \\
\bottomrule
\end{tabular}
}
\end{table}


\paragraph{Baselines.}
We compare HabitatAgent against six baselines representing common design choices:
\textbf{B1: Monolithic RAG}, a single-prompt dense-retrieval system without memory or verification;
\textbf{B2: Dense+Rerank}, a dense retriever (BGE) with a reranker, but without memory or verification;
\textbf{B3: GraphRAG-Fixed}, graph-based retrieval for all queries without adaptive routing;
\textbf{B4: LLM-Ranker}, which ranks a large candidate set without explicit verification;
\textbf{B5: Self-RAG}, which performs error detection followed by full regeneration; and
\textbf{B6: Rule-Verifier}, a rule-based verifier that refuses responses when errors are detected.
For a fair comparison, all methods use the same underlying LLM and, where applicable, the same candidate pool.

\subsubsection{Metrics}
We report metrics that reflect both recommendation utility and consultation reliability:
\begin{itemize}
    \item \textbf{Recommendation quality:} nDCG@5.
    \item \textbf{Constraint satisfaction:} \textbf{CSR@5} (Constraint Satisfaction Rate), the fraction of recommended items in top-5 that satisfy all \emph{hard} constraints.
    \item \textbf{Grounded generation quality:} RAGAS \textbf{Faithfulness}.
    \item \textbf{End-to-end accuracy (primary):} the percentage of responses that (i) satisfy all constraints, (ii) are factually correct with respect to retrieved evidence, and (iii) contain correct entity references.
    \item \textbf{System latency:} P95 end-to-end latency (ms).
\end{itemize}

\subsection{Main Results}
\label{subsec:main_results}

\paragraph{Overall performance (RQ1).}
Table~\ref{tab:main_results} reports results on all 300 queries. HabitatAgent achieves the best end-to-end accuracy (0.95) and the highest grounded generation faithfulness (0.96), outperforming all baselines. Compared with the strongest accuracy baseline, GraphRAG-Fixed (0.82), HabitatAgent improves end-to-end accuracy by 13 percentage points while reducing P95 latency by 100\,ms. This result suggests that reliable housing consultation requires more than stronger retrieval alone; it benefits from coupling persistent memory, evidence-grounded generation, and post-generation validation within a single workflow.

\paragraph{Complex queries and relational constraints (RQ1).}
Table~\ref{tab:complex_queries} reports results on the subset of 60 complex queries with at least three constraints. Dense+Rerank suffers a substantial drop in CSR@5 (0.08), indicating that semantic similarity alone is insufficient for hard relational constraints. HabitatAgent maintains 0.95 CSR@5 and 0.95 end-to-end accuracy, while remaining faster than always-on graph retrieval. This suggests that adaptive routing is effective in preserving relational correctness without incurring the full cost of graph-constrained retrieval on every query.

\begin{table}[t]
\centering
\caption{Performance on the complex query subset (60 queries with $\geq$3 constraints).}
\label{tab:complex_queries}
\resizebox{\columnwidth}{!}{
\begin{tabular}{@{}lccc@{}}
\toprule
\textbf{System} & \textbf{Accuracy} & \textbf{CSR@5} & \textbf{P95 Latency (ms)} \\
\midrule
B2: Dense+Rerank & 0.62 & 0.08 & 380 \\
B3: GraphRAG-Fixed & 0.85 & 0.88 & 820 \\
\textbf{Ours: HabitatAgent} & \textbf{0.95} & \textbf{0.95} & \textbf{680} \\
\bottomrule
\end{tabular}
}
\end{table}





\subsection{Ablation Study}
\label{subsec:ablation}

\subsubsection{Component efficacy (RQ2)}
We conduct ablations corresponding to the three methodological mechanisms described in \S\ref{sec:methodology}: \textbf{Verification-Gated Memory}, \textbf{Adaptive Retrieval Routing}, and \textbf{Failure-Type-Aware Remediation}. Table~\ref{tab:ablation} reports end-to-end accuracy after removing each component.

Removing \textbf{Adaptive Retrieval Routing} causes the largest degradation (0.95 $\rightarrow$ 0.75), indicating that selective graph-aware retrieval is critical for complex relational queries. This result suggests that graph-constrained retrieval should be invoked when query complexity requires it, rather than uniformly applied to all cases.

Removing the \textbf{Verification Gate} (0.95 $\rightarrow$ 0.88) and \textbf{Failure Remediation} (0.95 $\rightarrow$ 0.87) also leads to substantial drops, showing that reliability depends not only on retrieval quality but also on validated memory updates and recovery from validation failures. Finally, removing \textbf{Multi-Tier Validation} reduces accuracy to 0.85, highlighting the importance of explicit factual and entity checks before returning responses and updating memory.

\begin{table}[t]
\centering
\caption{Ablation study results (end-to-end accuracy).}
\label{tab:ablation}
\begin{tabular}{@{}lcc@{}}
\toprule
\textbf{Configuration} & \textbf{Accuracy} & \textbf{Change} \\
\midrule
Full System (HabitatAgent) & 0.95 & -- \\
\ \ \, w/o Verification Gate & 0.88 & $-7.0$pp \\
\ \ \, w/o Adaptive Routing & 0.75 & $-20.0$pp \\
\ \ \, w/o Failure Remediation & 0.87 & $-8.0$pp \\
\ \ \, w/o Multi-Tier Validation & 0.85 & $-10.0$pp \\
\bottomrule
\end{tabular}
\end{table}

\FloatBarrier

\section{Conclusion}

We presented HabitatAgent, a multi-agent architecture for end-to-end housing consultation. Rather than treating housing consultation as ranking or generation alone, HabitatAgent organizes memory, retrieval, generation, and validation into a closed-loop workflow for reliable decision support. Across 100 real consultation scenarios (300 multi-turn Q\&A pairs), the proposed design improves end-to-end accuracy from 75\% to 95\% over a strong Dense+Rerank baseline.

More broadly, our findings suggest that in high-stakes, multi-constraint decision-support tasks, correctness depends not only on retrieval quality or model capability in isolation, but also on how memory updates, evidence access, response generation, and validation are coordinated. Future work will extend the evaluation to more cities, larger datasets, and dynamically updated knowledge graphs.

\medskip
\small
\bibliographystyle{plain}
\bibliography{ref}

\appendix
\section{Product Demo}

 \vspace{-8mm}
\begin{figure}[ht]
  \centering
  \includegraphics[width=0.5\textwidth]{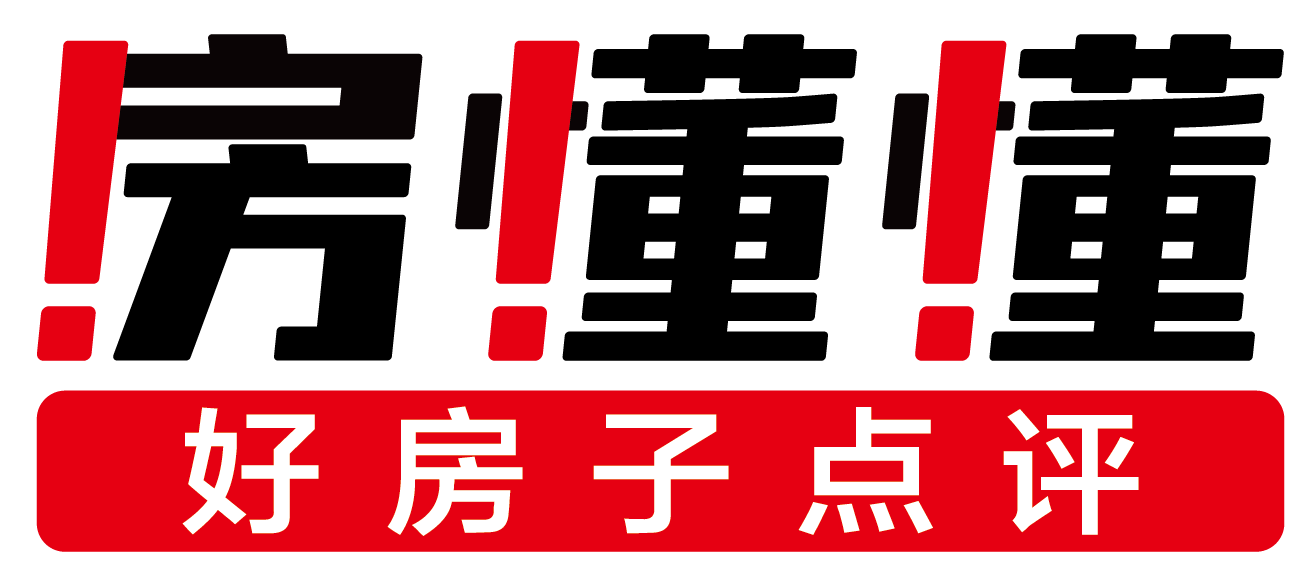} 
  \label{fig:FDD_logo}
\end{figure}

 \vspace{-8mm}
\begin{figure}[ht]
  \centering
  \includegraphics[width=0.5\textwidth]{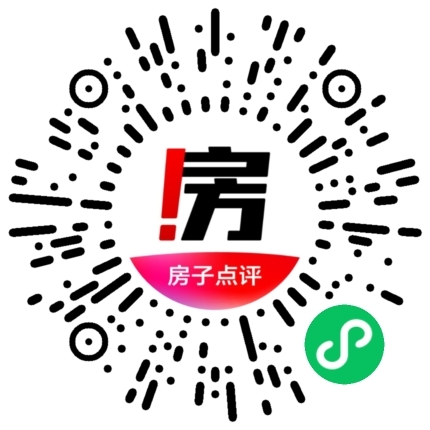} 
  \caption{Product Demo of Fangdongdong. Scan the QR code above, or search for "Fangdongdong" in WeChat to access the product and experience the real estate decision-making system powered by HabitatAgent.}
  \label{fig:product_demo}
\end{figure}

\end{document}